\documentclass[journal]{IEEEtran}
\usepackage{cite}
\usepackage{graphicx}
\usepackage{comment}
\usepackage{amsmath,amssymb}
\usepackage{color}
\usepackage{bm}
\usepackage{amsfonts}
\usepackage{array}
\usepackage{algorithm}
\usepackage{algorithmic}
\usepackage{cite}
\usepackage{booktabs}
\usepackage{float}
\usepackage{multirow}
\usepackage{mathrsfs}
\usepackage{amsfonts,amssymb} 
\usepackage{lipsum}
\usepackage{wrapfig}
\DeclareMathOperator*{\argmax}{arg\,max}

\usepackage{tabularx}
\usepackage{subfig}

\ifCLASSINFOpdf
\else
\fi
\begin{document}
\title{LTN: Long-Term Network for\\ Long-Term Motion Prediction}

\author{YingQiao Wang
\thanks{Y. Wang was with the Department
of Mathematics and Statistics, Colby College, Maine,
ME, 04901 USA e-mail: ywang22@colby.edu.}%

}

\markboth{Journal of \LaTeX\ Class Files,~Vol., No., October~2020}%
{Shell \MakeLowercase{\textit{et al.}}: Bare Demo of IEEEtran.cls for IEEE Journals}

\maketitle

\begin{abstract}
Making accurate motion prediction of surrounding agents such as pedestrians and vehicles is a critical task when robots are trying to perform autonomous navigation tasks. 
Recent research on multi-modal trajectory prediction, including regression and classification approaches, perform very well at short-term prediction. 
However, when it comes to long-term prediction, most Long Short-Term Memory (LSTM) based models tend to diverge far away from the ground truth. Therefore, in this work, we present a two-stage framework for long-term trajectory prediction, which is named as Long-Term Network (LTN). Our Long-Term Network integrates both the regression and classification approaches. We first generate a set of proposed trajectories with our proposed distribution using a Conditional Variational Autoencoder (CVAE), and then classify them with binary labels, and output the trajectories with the highest score. We demonstrate our Long-Term Network's performance with experiments on two real-world pedestrian datasets: ETH/UCY, Stanford Drone Dataset (SDD), and one challenging real-world driving forecasting dataset: nuScenes. The results show that our method outperforms multiple state-of-the-art approaches in long-term trajectory prediction in terms of accuracy.
\end{abstract}

\begin{IEEEkeywords}
Trajectory prediction, long short-term memory (LSTM), robots, autonomous vehicles
\end{IEEEkeywords}

\IEEEpeerreviewmaketitle

\section{Introduction}

\IEEEPARstart{A}{ccurately} predicting the motions of surrounding agents such as pedestrians and vehicles are significant when mobile robots or autonomous vehicles are trying to perform navigation tasks. In traffic systems, the future behavior of each traffic participant is determined by multiple aspects, such as the movement of other traffic agents, the physical constraints, and the traffic rules \cite{Jiachen_IROS19,salzmann2020trajectron,zhan2017safe}. 
Humans have the ability to navigate through a complex traffic scenario because they have the ability to reason about all the other people's actions, and how the physical constraints in the traffic systems affect their movements. Therefore, for a robot navigating through a complex traffic system, we need to consider about all the movement of the other surrounding traffic agents and the physical constraints in the traffic system.

With the discovery of the vanilla LSTM model, the researchers started to use Long Short-Term Memory networks to produce a regression of the future trajectories of traffic agents. The LSTM is a model that processes the data sequentially, so it is suitable for predicting the trajectories which is also considered as sequential data. 
Starting from the Social LSTM \cite{Alahi_2016_CVPR} model, the researchers started to model the people's social interaction. When predicting the future trajectories of traffic agents, they will store the knowledge about people, e.g. speed, direction, motion pattern, and people's social interaction in the hidden state \cite{Jiachen_ICRA19,ma2019wasserstein,Felsen_2018_ECCV,Jiachen_ITSC18-2}.

Then, the map information is integrated by extracting the features of map using a Convolutional Neural Network (CNN) combined with the current LSTM model, which largely improves the prediction accuracy. 
The recent works start to compete with each other by using different structures on modeling social interaction, and introduces a LSTM based encoder-decoder structure. The model encodes the past trajectory of the traffic agent using the LSTM along with the nearby traffic agents, produces a regression for the future trajectory, and decodes this trajectory using the LSTM.  
The Trajectron++\cite{salzmann2020trajectron}, WWTG\cite{Felsen_2018_ECCV} (Where Will They Go),and Social-BiGAT\cite{kosaraju2019socialbigat}, which are recent state-of-the-art models, outperform most of the popular LSTM model on future trajectory prediction in terms of accuracy. 
The model uses the traditional LSTM encoder-decoder structure, but it encodes the past trajectory and future trajectory into a latent space using the Conditional Variational Autoencoder (CVAE)\cite{NIPS2015_5775,Jiachen_social}. 
For prediction, it draws a latent variable from the latent space, decodes it as a regression using GRU\cite{cho2014learning} and past trajectory information. However, there is still space for improvement in terms of accuracy in long-term prediction, especially 3 to 4 seconds after the current observation.

In this work, we propose a two-stage framework called Long-Term Network (LTN) to improve the long-term trajectory prediction in terms of accuracy. In the first stage, the LTN uses a traditional LSTM-GRU encoder-decoder structure along with the CVAE\cite{NIPS2015_5775} to produce a set of possible future trajectory proposals. 
In the second stage, LTN performs classification and refinement on the trajectories proposals, and outputs the proposal with the highest score as the final trajectory prediction result. 
The trajectory proposals are generated based on the surrounding traffic agents identified by the LTN, and the prior extracted map information, so that the model can identify the traversable spaces of our robot and identify the possible effects of surrounding traffic agents to make better proposals. 

The contributions of this paper are summarized as follows: 1) We propose a newly modified GRU unit called Mogrifier GRU, based on the idea of the Mogrifier LSTM\cite{melis2020mogrifier}. By our refinement on the hidden state, we improve the performance of the model in terms of long-term prediction accuracy by 10\% just by replacing the regular GRU with our Mogrifier GRU. 2) We propose a two-stage approach, in which we combine the regression and classification methods and largely improve the performance on the long term trajectory prediction. 3) Our model achieve the state-of-the-art results on the widely used pedestrian trajectory prediction datasets (ETH/UCY)\cite{5459260}\cite{6909848}, Stanford Drone Dataset (SDD)\cite{10.1007/978-3-319-46484-8_33} and one real-world driving forecasting dataset (nuScenes)\cite{nuscenes2019}. 

\section{Related Work}

\subsection{Multi-Modal Trajectory Prediction }
Many earlier works in human trajectory forecasting can be roughly divided into two categories: the classic methods and the deep learning approaches. The classic methods include the kinematic equation, or the statistical models like polynomial fitting and Gaussian mixture models. 
In most recent works, Recurrent Neural Network (RNN) and its variants such as Long Short-Term Memory (LSTM) or Gated Recurrent Unit (GRU)\cite{cho2014learning}, and Convolutional Neural Network become the basis of most of recent models. Researchers utilize CNN to extract map features, and RNN or its variants, LSTM and GRU, to capture the social interaction between the traffic agents, and then regress the future trajectory \cite{li2019coordination,kosaraju2019socialbigat,fang2020tpnet,li2018generic,Jiachen_TITS}.

\subsection{Deep Generative Models}
Besides the classic methods and the CNN-RNN combined approaches, the generative approaches have emerged as another state-of-the-art approach in trajectory prediction problems. 
This approach shifts some researchers' focus on regressing a single future trajectory to producing a distribution of the future trajectory. 
With a full distribution of the future trajectory, the results can produce more possible trajectory proposals. 
To produce such distributions, most works use recurrent backbone architecture with a latent variable model, such as the Generative Adversarial Network (GAN)\cite{goodfellow2014generative}, and the Conditional Variational Autoencoder (CVAE)\cite{NIPS2015_5775}. 
Currently, Trajectron++\cite{salzmann2020trajectron}, Social-BiGAT\cite{kosaraju2019socialbigat}, WWTG\cite{Felsen_2018_ECCV} are two CVAE and GAN based models that outperform most state-of-the-art trajectory prediction models. Trajectron++\cite{salzmann2020trajectron} and Social-BiGAT \cite{kosaraju2019socialbigat} are able to account for the social interactions between traffic agents and physical constraints in the scene.

\subsection{Regression and Classification}
Current models that produce full distributions mostly utilize the Gaussian Mixture Model, which outputs the local maximum of the distribution as the final trajectory prediction result. 
But empirically, by the qualitative analysis in most of the work, the output is not actually the closest trajectory produced by the full distribution to the ground truth.
So the new approaches combining regression and classification appear, which generates a set of hypothesis trajectory proposals, and outputs the proposal with the highest score as the final trajectory prediction result. Trajectory Proposal Net (TPNet)\cite{fang2020tpnet} is another state-of-the-art that uses this regression and classification method for trajectory prediction, where the model does polynomial fitting between the starting point and the proposed end point of the traffic agent, while considering the social-interaction and traffic rules. 
At the end, the model performs classification on these proposed trajectories and outputs the proposal with the top scores. 

\section{Problem Formulation}
In this work, we select our robot as the center point in the scene, where we will determine its surrounding traffic agents and predict their future trajectories. 
During the time interval $[0,T_{obs}] \cup [T_{obs}+1,T_{future}] $, we denote the number of traffic agents surrounding the robot at time $t \in [0,T_{obs}] \cup [T_{obs} +1,T_{future}]$ as $S$. 
We denote the surrounding agents as $\{A_1 .... A_S\}$, and for each agent, we categorize it as pedestrians or vehicles. 
For simplicity, the vehicle category also includes agents like bicycles, motorcycles, and cars. 
The model takes a series of past positions in the time interval $[0,T_{obs}]$ of each agent $P_{A_i}^{obs}=\{P_{A_i}^{0},...,P_{A_i}^{T_{obs}}\}$, and also a series of past positions of the robot $P_r^{obs}=\{P_r^{0},...,P_r^{T_{obs}}\}$, where $i \in [0,S]$. For each agent's future trajectory in interval $[T_{obs}+1,T_{future}]$, we denote it as $P_{A_i}^{future}=\{P_{A_i}^{T_{obs}+1}...P_{A_i}^{T_{future}}\}$, and for the robot's future trajectory, we expresses it as $P_r^{future}=\{P_r^{T_{obs}+1}...P_r^{T_{future}}\}$, where $i \in [1,S]$.
We also incorporate the map information in the same way as Trajectron++, and encode the map information as $M=M_S^t$, where the $S$ is the agent, and $t \in [0,T_{obs}] \cup [T_{obs}+1,T_{future}]$.

\section{Method}

\subsection{Long-Term Network}
To further improve the performance of current model in long-term trajectory prediction, we propose a two-stage framework called Long-Term Network (LPN). The framework is visualized in Figure 1. 

\subsection{Determining the Surrounding Traffic Agents}
To determine the surrounding traffic agents, we first determine the number of agents in the scene, and denote it as $S$. We include the agents $A_i$ ($i \in [1,S]$) that are close to the robot in $\ell_2$ distance. 
Formally, the agent is selected if at time $t \in [0,T_{obs}] \cup [T_{obs} +1,T_{future}]$, $\|{P_{A_i}^t-P_r^t}\|^2 \leq d$, where $d$ is a hyperparameter indicating the maximum perception distance. 

Since we are going to predict each agent $A_i$'s future trajectories, we perform the similar process to select the surrounding traffic agents of our selected agents $A_i$. The agents around $A_i$ is again determined by the $\ell_2$ distance. 
Formally, the agent around $A_i$ is selected if at time $t \in [0,T_{obs}] \cup [T_{obs} +1,T_{future}]$ and $x \in [0,S]_{\ne{i}}$, $\|{P_{A_i}^t-P_{A_x}^t}\|^2 \leq d$, where $d$ again is the same hyperparameter that expresses the maximum perception distance.

\begin{figure*}[t]
 \centering
 \includegraphics[scale=0.18]{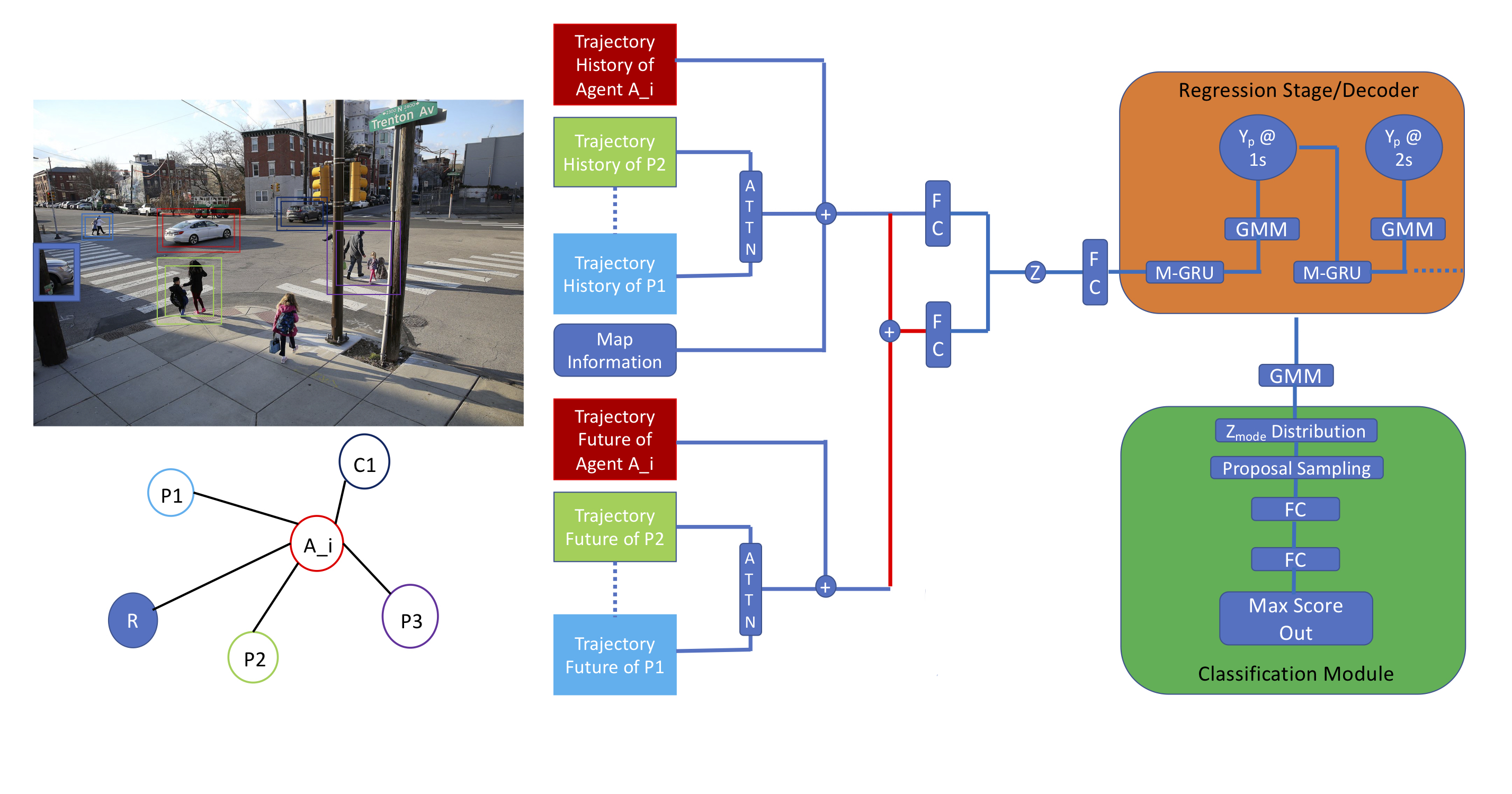}
 \
 \caption{Our visualization to the LTN model. The left is a small graph visualization how our model determines the surrounding agents of agent $A_i$ at a real-life complex traffic intersection. The right is the model structure with each module indicated out.}
 \label{Figure 1}
\end{figure*}

\subsection{Modeling the Agent History and The Social Interactions}
To model the agent history, we primarily utilize the Mogrifier LSTM\cite{melis2020mogrifier}, which is a variant of the vinilla LSTM model. 
The Mogrifier LSTM has better performance in long-term performance than the vinilla version, as the experiment in the paper demonstrates. 
The Mogrifier LSTM utilizes the same LSTM module, but between each unit, the Mogrifier LSTM updates the input and previous hidden state with several rounds of mutual gating, which is called a mogrifying step. 
The Mogrifier LSTM can also be implemented based on Bi-directional LSTM. Since there are no public code for the Mogrifier LSTM, we implement the bi-directional Mogrifier LSTM with traditional bi-directional LSTM and add Mogrifier into the model. 
The number of mogrifying steps is 6, and the number of layers is 2. 
To model the agent history, we input the agent $A_i$'s history trajectory $P_{A_i}^{obs}=\{P_{A_i}^{0}...P_{A_i}^{T_{obs}}\}$ into a Mogrifier LSTM network with 32 hidden dimensions to obtain the encoded agent history tensor.

To model the social interactions, we input the agent $A_i$'s surrounding traffic agents trajectories $P_{A_x}^{obs}=\{P_{A_x}^{0}...P_{A_x}^{T_{obs}}\}$, which $x \in [0,S]_{\ne{i}}$, into a Mogrified LSTM with 8 hidden dimensions. 
Then, we utilize an attention module, which encodes these social interactions as additive attentions. 
We utilize additive attention, where the encoded tensors of all surrounding traffic agents $A_x$ are aggregated to obtain one attention tensors, and then concatenate with the corresponding agent history tensors to obtain one complete history tensor $V_i$.

\subsection{Map Encoding and Future Encoding} 
To obtain the encoded map information, we utilize Convolutional Neural Network (CNN) to encode the local map information, which is the similar to Trajectron++\cite{salzmann2020trajectron}. 

We model the target agent and its surrounding traffic agents' future trajectories into the encoded tensors during the training phase, in order to provide information to formulate the future trajectory distribution used in the training phase. 
We input the agent $A_i$'s future trajectory $P_{A_i}^{future}=\{P_{A_i}^{T_{obs}+1}...P_{A_i}^{T_{future}}\}$ into a Mogrifier LSTM network with 32 hidden dimensions to obtain the encoded agent future tensor, and we input the agent $A_i$'s surrounding traffic agents trajectories $P_{A_x}^{future}=\{P_{A_x}^{T_{obs}+1}...P_{A_x}^{T_{future}}\}$, which $x \in [0,S]_{\ne{i}}$ into the Mogrifier LSTM with 8 hidden dimensions. Then, the additive attention is used to aggregate both tensors to obtain one attention tensor, and then we will concatenate our attention tensor into the corresponding agent's future tensor to obtain one complete future tensor $V_f$.

\subsection{CVAE Latent Variable Framework}
To address the multi-modality to produce the full distribution of the agent $A_i$'s future trajectory, we utilize the modified Conditional Variational Autoencoder (CVAE)\cite{NIPS2015_5775} latent variable framework employed in \cite{IvanovicLeungEtAl2020} and \cite{8500493}. 
The framework utilizes our encoded complete history tensor $V_i$ and produces the full distribution $p(V_f|V_i)$ on the future trajectory of agent $A_i$ by defining a discrete categorical latent variable $z \in Z$ which we then can express $p(V_f|V_i)$ as:
\begin{equation}
    p(V_f|V_i) = \sum_{z \in Z} p_\psi(V_f|V_i,z,M)p_\theta(z|V_i,M)
\end{equation}
where $|Z|=25$ and $\psi,\theta$ are neural network weights that parameterize their respective distributions, and $z$ is a discrete latent variable that also aids in interpretability. 

During training, we have $V_f$ in the dataset and use a bi-directional Mogrifier LSTM with 32 hidden dimensions to produce a ground truth future trajectory distribution $q_{\phi}(z|V_f,V_i,M)$, and $\phi$ is neural network weight again.

\subsection{Mogrifier GRU}
To obtain the final trajectory distribution, we need to decode the final trajectory distribution with the latent variable $z$, and the complete history tensor $V_i$, using the GRU\cite{cho2014learning} units. 
We present a more powerful variant of GRU unit, where we adopt the similar idea in Mogrifier LSTM\cite{melis2020mogrifier} to process the input and hidden state before each GRU unit. 
Suppose we have input $x_{input}$ and previous hidden state $h_{prev}$, for a normal GRU, the current hidden state $h_cur$ is calculated by:
\begin{equation}
    h_{cur}=GRU(x_{input},h_{prev}),
\end{equation}
and the $h_{cur}$ is calculated by:
\begin{equation}
    \begin{split}
       r_{cur}=&\sigma(W_{ir}x_{input}+b_{ir}+W_{hr}h_{prev}+b_{hr})\\
       z_{cur}=&\sigma(W_{iz}x_{input}+b_{iz}+W_{hz}h_{prev}+b_{hz})\\
       n_{cur}=&\tanh(W_{in}x_{input}+b_{in}+r_{cur}*(W_{hn}h_{prev}+b_{hn})\\
       h_{cur}=&(1-z_{cur})*n_{cur}+z_{cur}*h_{prev},
    \end{split}
\end{equation}
where $r_{cur},z_{cur},n_{cur}$ are the reset, update, and new gates. $\sigma$ is the sigmoid function, and $*$ is the Hadamard product, and all the $W,b$ are the learnable weights matrices.

Our Mogrifier GRU works by performing mogrifying steps before the usual GRU computation step. Suppose we perform mogrifying steps $i$ times, we have MogGRU$(x_{input},h_{prev})$ = GRU$(x_{input}^{i},h_{prev}^{i})$. 
For each $a \in [1,i]$:
\begin{equation}
    x_{input}^a=1.5\tanh(Q^ah_{prev}^{a-1})\odot x_{input}^{a-2},
\end{equation}
\begin{equation}
    h_{prev}^a=1.5\tanh(R^ax_{input}^{a-1})\odot h_{prev}^{a-2},
\end{equation}
where for odd $a \in [1,i]$, the equation $(1)$ is performed, and for even $a \in [1,i]$, the equation $(2)$ is performed. The parameters in the equations are recommended as: $i \in [5,6]$ and $i$ has to be a integer. In our model, we choose $i=6$, and $i=0$ will recover the traditional GRU unit. The parameter $Q,R$ in the mogrifying steps are the same as the $Q,R$ in the parameters in the Mogrifier LSTM, which are randomly initialized matrices. We will have the experiment result of Mogrifier GRU in the experiment section to show it's performance comparing with the traditional GRU. 

\subsection{Trajectory Proposal and Classification Module}
During training, when we get the full distribution of the future trajectory of agent $A_i$, we then produce the final trajectory proposal by sampling $N$ numbers of the trajectories from the latent variable distribution, where the latent variable is determined by:
\begin{equation}
    z=\argmax_{z \in Z} p_\theta (z|V_i,M).
\end{equation}
Then, in the classification module. We consider the methods used in TPNet\cite{fang2020tpnet}. We assign each proposal with a binary class label, which is used to indicate whether it is a good trajectory or not. 
We use the average distance between all the sampled trajectory proposals and the ground truth proposal as the criterion for measuring the proposal's quality, which can be expressed as:
\begin{equation}
    D = \frac{1}{N}\sum^{N}_{n=1} \|{p_{A_i^{gt}}^{n}-p_{A_i^{prop}}^{n}}\|^2,
\end{equation}
which is the average $\ell_2$ distance between the n-th sampled proposal vector $p_{A_i^{prop}}^{n}$ and the ground truth trajectory $p_{A_i^{gt}}^{n}$. Then, a threshold $\gamma$ is used, and for the proposed trajectories that have $D$ lower than $\gamma$, we assign positive labels. For the proposed trajectories that have $D$ larger than $\gamma$, we assign negative labels. 

\subsection{Objective Function}
In our model, there are two loss functions to be minimized, One is the regression loss, and the other one is the classification loss. 

\textbf{Regression Loss:}
We adopt the objective function provided in Trajectron++\cite{salzmann2020trajectron} and WWTG\cite{Felsen_2018_ECCV}, Where we aim to solve:
\begin{equation}
    \begin{split}
    &L_{reg}=\max_{\phi,\theta,\psi} \sum_{r=1}^{N} \mathbb{E}_{z\sim q_{\phi}(z|V_{i_r},V_{f_r},M)}[\log p_{\psi}(V_{f_r}|V_{i_r},z,M)]\\
    &-\beta {D_{\mathrm{KL}}(q_{\phi}(z|V_{i_r},V_{f_r},M)||p_{\theta}(z|V_{i_r},M))}
    +\alpha I_q(V_{i_r};z),
    \end{split}
\end{equation}
where by \cite{salzmann2020trajectron}, the $I_q$ is the mutual information between $V_i$ and $z$ under the distribution $q_{\phi}(V_i,z)$. 
The computation of $I_q$ is the same as \cite{article}. $\alpha$ and $\beta$ are hyperparameters.

\textbf{Classification Loss:}
For the classification loss, we consider the methods in TPNet\cite{fang2020tpnet}, which a binary cross-entropy loss $L_{class}$ is employed as:
\begin{equation}
    L_{class}(x,y)= -w(y*log(x)+(1-y)*log(1-x))
\end{equation}

The total loss is written as follows: 
\begin{equation}
    L_{total} = L_{reg}+ \frac{1}{N} \sum_{n}^{N }L_{class}(c_n,c_n^{*})
\end{equation}
where $N$ is the number of trajectory proposals in the proposal set, $w$ is learnable weight, and $c_n^{*}$ is the corresponding predicted label, and $c_n$ is the corresponding ground truth label. 

During the training phase, the regression module minimizes the regression loss, and the classification module minimizes the classification loss.

\section{Experiments}

\subsection{Datasets}
Our model is evaluated on four widely used public datasets: The ETH, UCY, Stanford Drone Dataset, and nuScenes. The ETH and UCY datasets foucs on the pedestrian trajectory prediction, and contains complex social interactions. The ETH/UCY dataset has five subsets, each named ETH, HOTEL, UCY, ZARA-01, ZARA-02.  There are two settings for the length of trajectories, $T_{obs}=T_{future}=3.2s$ and $T_{obs}=3.2s$, $T_{future} =4.8s$. The data is captured at $2.5 Hz$ ($\Delta t=0.4s$), 
So the dataset will contains 8 frames for observations and 8/12 frames for prediction.

For the Stanford Drone Dataset, this is a trajectory dataset that is captured by drones from top-down view. So the scenes in the dataset are top-down-view. The scene are captured at a university campus with vehicles, cyclists, and crowds. The dataset contains a lot of heterogeneoous data.

For the nuScenes dataset, this is a challenging large real-world driving forecasting dataset, where with more than 1000 scenes in the dataset are captured in Boston and Singapore. Each scene is 20 seconds long, and the dataset contains High-Definition semantic maps. All the scenes in the dataset contains a large amount of heterogeneous data, with complex social interactions among up to 23 semantic object classes. Also, the map provides data about the physical constraints in each scenes. 

\subsection{Evaluation Metrics}
In our experiment, we will use four metrics that are commonly used in all trajectory prediction models, which include $\text{minADE}_{20}$, $\text{minFDE}_{20}$, $\text{ADE}_{20}$ and $\text{FDE}_{20}$.
\begin{itemize}
    \item $\text{minADE}_{20}$ ($\text{mADE}_{20}$): the minimum mean $\ell_2$ distance between the ground truth trajectory and predicted trajectory from the best of 20 samples by the LTN.
    \item $\text{minFDE}_{20}$ ($\text{mFDE}_{20}$): the minimum $\ell_2$ distance between the ground truth final position and the predicted final position at the final ${T_{future}}$ from the best of 20 samples by the LTN.
    \item $\text{ADE}$: the mean $\ell_2$ distance between the ground truth trajectory and predicted trajectory by the LTN.
    \item $\text{FDE}$: the $\ell_2$ distance between the ground truth final position and the predicted final position at the final ${T_{future}}$ by the LTN.
\end{itemize}

\begin{table*}
\begin{center}
\caption{Comparison between Mogrifier GRU version LTN (LTN$_{M20}$) and the baseline methods on ETH and UCY benchmark. We used $T_{future}=4.8$ seconds setting. Each row represents a dataset and each column represents a method, with ADE and FDE separated into two tables for better clarity. The ADE and FDE are measured in Meter. All the $_{20}$ marks means that the prediction is chosen from the best of 20 samples.}
\begin{tabular}{|c|c|c|c|c|c|c|c|c|c|c|c|} 
\hline
Metric & Dataset & LSTM & S-LSTM & S-GAN$_{20}$ & Trajectron++$_{20}$ & S-BiGAT$_{20}$ & TPNet$_{20}$ & SoPhie$_{20}$ & STGAT & LTN$_{M20}$ \\
\hline
\multirow{5}{2em}{ADE} & ETH & 1.09 & 1.09 & 0.81 & 0.43 & 0.69 & 0.84 & 0.70 & 0.65 & \textbf{0.39}\\ 
& HOTEL & 0.86 &0.79 &0.81 &\textbf{0.12} &0.69 &0.24 &0.76 &0.49 &0.16\\ 
& UNIV & 0.61 &0.67 &0.72 &0.22 &0.4 &0.42 &0.54 &0.55 &\textbf{0.20}\\ 
& ZARA1 & 0.41 &0.47 &0.60 &0.17 &0.55 &0.33 &0.30 &0.30 & 0.18\\
& ZARA2 & 0.52 &0.56 &0.34 &\textbf{0.12} &0.30 &0.26 &0.38 &0.36 &0.15\\
& \textbf{Average} & 0.70 &0.72 &0.42 &\textbf{0.20} &0.36 &0.42 &0.54 &0.48 &0.22\\
\hline
\end{tabular}
\begin{tabular}{|c|c|c|c|c|c|c|c|c|c|c|c|} 
\hline
Metric & Dataset & LSTM & S-LSTM & S-GAN$_{20}$ & Trajectron++$_{20}$ & S-BiGAT$_{20}$ & TPNet$_{20}$ & SoPhie$_{20}$ & STGAT & LTN$_{M20}$ \\
\hline
\multirow{5}{2em}{FDE} & ETH & 2.41 & 2.35 & 1.52 & 0.86 & 1.29 & 1.73 & 1.43 & 1.12 & \textbf{0.80}\\ 
& HOTEL & 1.91 &1.76 &1.61 &0.19 &1.01 &0.46 &1.67 &0.66 &\textbf{0.18}\\ 
& UNIV & 1.31 &1.40 &1.26 &0.43 &1.32 &0.94 &1.24 &1.10 &\textbf{0.41}\\ 
& ZARA1 & 0.88 &1.00 &0.63 &0.32 &0.62 &0.75 &0.63 &0.69 & \textbf{0.29}\\
& ZARA2 & 1.11 &1.17 &0.84 &0.25 &0.75 &0.60 &0.78 &0.60 &\textbf{0.23}\\
& \textbf{Average} & 1.52 &1.54 &1.18 &0.41 &1.00 &0.90 &1.15 &1.08 &\textbf{0.38}\\
\hline
\end{tabular}
\label{table:1}
\vspace{0.5cm}
\caption{Comparison between Mogrifier GRU version LTN (LTN$_{M20}$) and the baseline methods on SDD benchmark. Each row represents a dataset and each column represents a method, with miniADE and miniFDE separated into two tables for better clarity. All the miniADE and miniFDE are measured in Pixels and are the measurement for predictions at $T_{future}=4.8$ seconds. All the $_{20}$ marks means that the prediction is chosen from the best of 20 samples.}
\begin{tabular}{|c|c|c|c|c|c|c|c|c|} 
\hline
Metric & Dataset & S-LSTM & S-GAN & S-ATTN & STGAT & Trajectron++ & CAR-Net & LTN$_{M20}$ \\
\hline
\multirow{1}{4em}{mADE$_{20}$} & SDD & 31.4 & 27.0 & 33.3 & 18.8 & 19.3 & 25.72 & \textbf{15.2} \\ 
\hline
\end{tabular}
\begin{tabular}{|c|c|c|c|c|c|c|c|c|} 
\hline
Metric & Dataset & S-LSTM & S-GAN & S-ATTN & STGAT & Trajectron++ & CAR-Net & LTN$_{M20}$ \\
\hline
\multirow{1}{4em}{mFDE$_{20}$} & SDD & 55.6 & 43.9 & 55.9 & 31.3 & 32.7 & 51.80 & \textbf{25.8} \\ 
\hline
\end{tabular}
\label{table:2}

\end{center}
\end{table*}

\subsection{Baselines}
We compare the performance of our model with the state-of-the-art models below:
\begin{itemize}
     \item Vanilla LSTM: An LSTM network utilizing only the agent
     $A_i$'s history trajectory information. 

\item Social LSTM\cite{Alahi_2016_CVPR}: An LSTM network that utilizes not only the agent $A_i$'s history trajectory information, but also uses LSTM to model the agents' trajectory information around agent $A_i$. 

\item Social GAN\cite{gupta2018social} (S-GAN): This is a GAN with social interaction considered. Each agent is modeled by an LSTM-GAN combined network, where the LSTM encoder-decoder outputs are the generator of GAN, and the generated trajectories are then evaluated against the ground truth trajectories in the discriminator. 

\item Trajectron++\cite{salzmann2020trajectron}: This is a CVAE-based trajectory prediction model, where the LSTM-GRU encoder-decoder structure is used, and the LSTM is used to model the agent's history and it's corresponding social interaction, and at the GRU decoder a full distribution of the predicted trajectory is produced. 

\item Social-BiGAT\cite{kosaraju2019socialbigat} (S-BiGAT): This is an LSTM-GAN with Graph Attention Network to encode agent's social interactions. 

\item TPNet\cite{fang2020tpnet}: This is a CNN based network that produces a trajectory proposal set by first predict the end point from the given map information and then predict the potential endpoint, then do regression based on map information, starting and the end point. At the end, a classification is performed to output proposals with high scores. 

\item Sophie\cite{8953374}: This is a GAN-based trajectory prediction model that also leverages the social interactions and physical information. Similar to Social-GAN, the trajectory is produced by generator and the discriminator will evaluate these predictions against the ground truth trajectories. 

\item STGAT\cite{Huang_2019_ICCV}: This is a spatial-temporal graph based trajectory prediction model. The spatial interaction is captured by graph attention mechanisms and the LSTM is used for temporal interactions.

\item Social-Attention\cite{vemula2018social}(S-ATTN): This is an attention based trajectory prediction model, where it captures the relative importance, which is attention, of each person when predicting for the future trajectories.

\item Car-Net\cite{sadeghian2018carnet}: This is a prediction model that can account the dependencies between agent's behavior and their spatial environment, where the model can learns where to look in a large environment when predicting the trajectory of an agent.

\item CSP\cite{8500493}: This is a prediction model that builds on LSTM encoder-decoder framework, where LSTM is used to model the agent's social interaction and output multi-modal future distribution of the agent based on the social interaction.

\item SpAGNN\cite{casas2019spatiallyaware}: This is a probabilistic model that utilizes graph neural network to capture the interactions between the vehicles and output a distribution of the future trajectory of the vehicle the model selected to predict.
\end{itemize}

\textbf{Implementation Details}
For our experiment, we adopted the ETH/UCY/nuScenes dataset preprocessing method provided in Trajectron++ \cite{salzmann2020trajectron}. For SDD, it is preprocessed according to the methods provided in Evolvegraph \cite{li2020Evolvegraph}. For our Mogrifier GRU, we chose the mogrifying step to be 6, and the $\alpha$, $\beta$ in the regression objective function is $\alpha$=1, and $\beta$ is dynamically changed for the best performance, according to Trajectron++, which this methods provides most optimal result. We optimize the network using Adam\cite{kingma2017adam} optimizer with learning rate 0.002. The $\gamma$ we used is 3$m$. We implemented LTN with PyTorch on a desktop with Ubuntu 20.04, equipped with one Intel I7-8700K CPU and two Nvidia GTX 1070Ti GPUs.

\subsection{Evaluation of Trajectory Prediction}
We compared our Mogrifier GRU version LTN with several baseline method on ETH, UCY, datasets in terms of two metrics ADE and FDE in Table I.  We reported our results as LTN$_{M20}$, where it's the 20 trajectory proposal that has the highest scores. 
In Table I, we can see that in terms of ADE, our LTN$_{M20}$ can outperform Trajectron++, which leads the rest of the model in the data we collected. 
In ETH and UNIV, our model achieved better ADE results. 
Also, in terms of FDE, our LTN$_{M20}$ outperforms all other methods completely, showing that our model is very good at long term trajectory prediction. 
We successfully improved our methods by nearly 10\% in terms of FDE, or long term prediction. However, we also notice that both ETH and UCY has reached a saturation, which means that more improvement is unlikely due to the data annotation errors or any off-errors during data collection, which leads us to analyze the LTN's performance in another dataset SDD.

We compared our Mogrifier GRU version LTN with several baseline method on SDD dataset in terms of minADE$_{20}$ and minFDE$_{20}$ in table 2. In table 2, our methods outperformed all the baseline methods, with over $20$ percent improvement in both minADE$_{20}$ and minFDE$_{20}$. 

Not only our LTN demonstrated its long-term prediction advantages in terms of accuracy in ETH, UCY, and SDD, we also introduced a dataset with more heterogeneous data, nuScenes.

\begin{table}
\centering
\caption{Comparison between Mogrifier GRU version LTN (LTN$_{M20}$) and the baseline methods on nuScenes benchmark. We predicted the trajectories $4$ seconds after the $T_{obs}$. Each row represents methods, and each column represents the FDE of each methods at that discrete time period. All the FDE are measured in Meters.}
\begin{tabular}{ |c|c|c|c|c|} 
\hline
Methods & 1s & 2s & 3s &4s\\
\hline
 S-LSTM & 0.47 & - & 1.61 & -\\ 
 CSP & 0.46 &- &1.50 &- \\ 
 CAR-Net & 0.38 &-&1.35 &- \\ 
 SpAGGN & 0.36 &- &1.23 &- \\
 Trajectron++ & \textbf{0.07} &0.45 &1.14 &2.20 \\
 LTN$_{M20}$ & 0.13 &\textbf{0.43} &\textbf{0.92} &\textbf{1.74} \\
\hline
\end{tabular}
\label{table:3}
\end{table}

In Table III, we compared our LTN$_M20$ model with several baseline  methods  with  nuScenes dataset.  We  can  see  that Trajectron++  outperformed  all  the  other  baseline  methods across all 4 seconds time span, but our LTN$_{M20}$ outperforms Trajectron++, especially in long terms, where over 20 percent improvement  is  achieved  in3seconds  and4seconds.  But we  also  noticed  that  in  short  term,  especially  in1seconds,our  method  did  not  outperform  Trajectron++,  but  we  are  not especially concerned with it because our model demonstrated persuasive   long   term   prediction   performance   in   terms   of accuracy.

\subsection{Evaluation of Mogrifier GRU}
We now examine the performance of the Mogrifier GRU. As we mentioned before, because we believe both ETH/UCY datasets have reached a saturation, where there are no possible improvement space for us, we conduct the rest of the experiment on SDD and nuScenes datasets. As the table shows, we set up experiment with two versions of Trajectron++, since the authors kindly provided their code:

Version 1: Trajectron++ with regular GRU as decoder.

Version 2: Trajectron++ with Mogrifier GRU as decoder.

\begin{table}
\centering
\caption{Comparison between Mogrifier GRU version Trajectron++ (Trajectron++$_{M}$) and Regular GRU version Trajectron++ on nuScenes benchmark. We predicted the trajectories $4$ seconds after the $T_{obs}$. All the FDE are measured in Meters.}
\begin{tabular}{|c|c|c|c|c|} 
\hline
Methods & 1s & 2s & 3s &4s\\
\hline
 Trajectron++ & \textbf{0.07} &0.45 &1.14 &2.20 \\
 Trajectron++$_{M}$ & 0.08 &  \textbf{0.43} & \textbf{1.02} & \textbf{2.03}\\
\hline
\end{tabular}
\label{table:4}
\end{table}

As Table IV shows, the Mogrifier GRU version Trajectron++ outperforms the regular GRU version Trajectron++ at $3$ seconds and $4$ seconds. At $3$ seconds and $4$ seconds the improvement reached around $10$ percent just by switching the regular GRU with our Mogrifier GRU. Our Mogrifier GRU is intended to remind the future hidden states and input with the information before, so our experiment showed that with the reminder of the states of the agent in $1$ seconds and $2$ seconds provided to $3$ seconds and $4$ seconds time period, the prediction result in $3$ seconds and $4$ seconds improved.

\begin{figure*} 
    \centering
  \subfloat[\label{2a}]{%
       \includegraphics[width=0.45\linewidth]{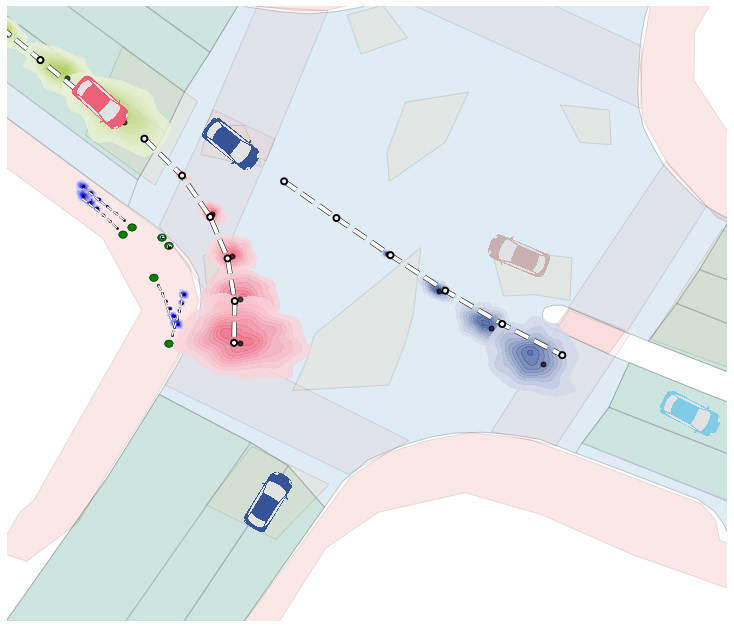}}
    \hfill
  \subfloat[\label{2b}]{%
        \includegraphics[width=0.45\linewidth]{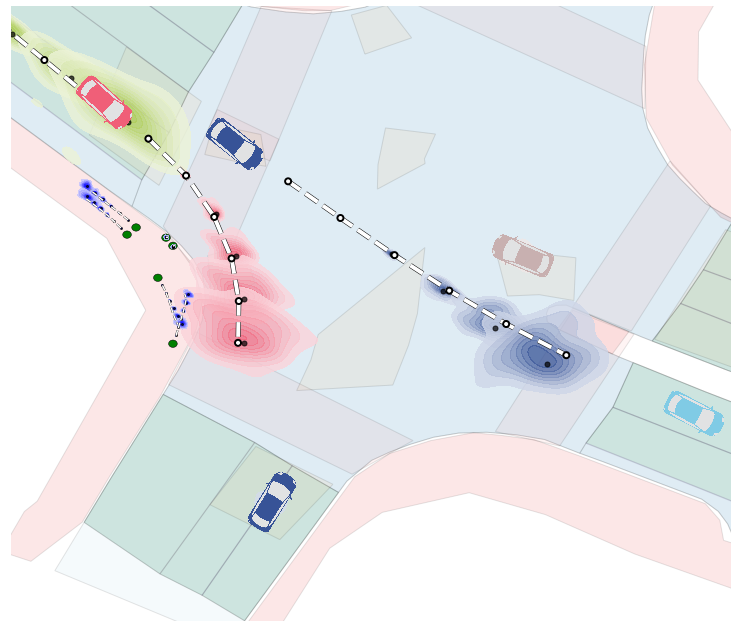}}
  \subfloat[\label{2c}]{%
        \includegraphics[width=0.45\linewidth]{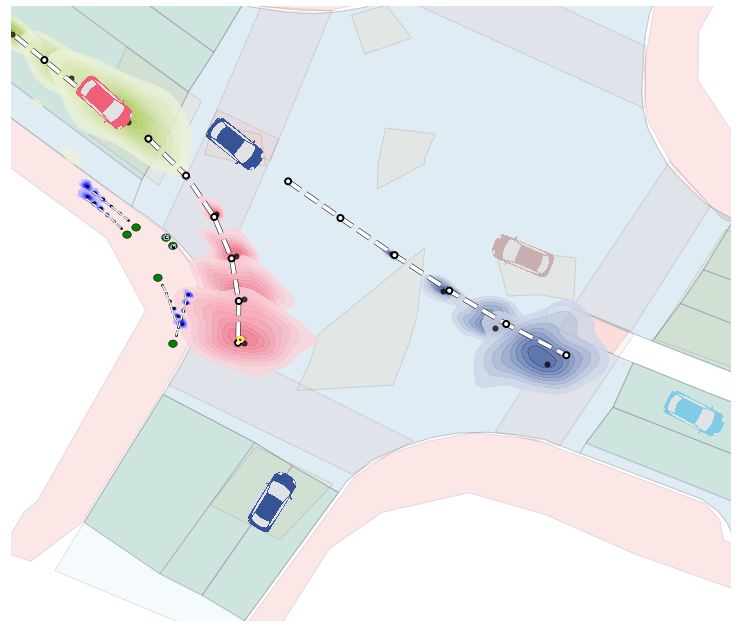}}
    \hfill
  \caption{The same scene with predictions from Trajectron++\cite{salzmann2020trajectron} and LTN$_{M20}$. (a) The prediction from Trajectron++ where a most likely single output and the full distribution of future trajectory are provided. (b) The $z_{mode}$ distribution prediction from LTN$_{M20}$ and the most likely single output from the Trajectron++ are provided. (c) The prediction of $T_{future}=3$ seconds from LTN$_{M20}$, denoted in a yellow circle, and the $z_{mode}$ distribution are provided.}
  \label{Figure 2} 
\end{figure*}

\subsection{Qualitative Analysis}

We compared our visualized experiment results with Trajectron++\cite{salzmann2020trajectron} in the same scene provided in nuScenes\cite{nuscenes2019} dataset. 
In Fig. 2(a) and Fig. 2(b), we can clearly see the advantage brought by the $z_{mode}$ distribution. We can compare the shape of the distribution in both figures, where we can see that $z_{mode}$ distribution of the red and blue car looks more uniform, meaning that it can generate samples with less bias. Therefore, the $z_{mode}$ distribution in Fig. 2(b) can provide more samples near the ground truth. 
Also, the dense center of the $z_{mode}$ distribution shifts towards the ground truth comparing with that of the full distribution. So, the proposed samples in $z_{mode}$ will have larger probability to be the closest samples to the ground truth.
Although the variance increases for the $z_{mode}$ distribution, we will tackle it with our classification module, where the samples with high variance will be filtered out by the threshold $\gamma$. 

In Fig. 2(b), we can also see that by plainly taking the $\text{argmax}$ of the $z_{mode}$ distribution, the output is not the closest one to the ground truth. This is very clear in Fig. 2(b), where for the red car and its end point prediction, the $z_{mode}$ indicates an area at the very center of the distribution that is closest to the ground truth trajectory, but the most likely output does not fall into that area. 
By our classification module, with good sampling from the $z_{mode}$ distribution, filtering, and classification, we obtained the yellow hollow circle in Fig. 2(c) that falls into the area indicated by the $z_{mode}$ distribution. The result is better compared with Trajectron++. For clarity, we only indicate the prediction of LTN$_{M20}$ at $T_{future}=3$ seconds, which is the end point distribution in red for the red car.

\section{Conclusion}
In this work, we present our model LTN, a two-stage trajectory prediction model for long-term trajectory prediction. The LTN incorporates heterogeneous data and combines regression and classification methods to improve the trajectory prediction performance for long-term prediction. Our LTN will first generate a distribution of the future trajectories, sample future trajectory proposals from it and perform classification on our proposal set. Along with the data of surrounding agents and the map information, our model could ensure that the final trajectory prediction is ideal and better. Our LTN achieves state-of-the-art performance in various popular datasets and shows significant improvement in terms of long-term trajectory prediction accuracy.

\ifCLASSOPTIONcaptionsoff
  \newpage
\fi

\bibliographystyle{IEEEtran}
\bibliography{IEEEabrv,reference}

\end{document}